\documentclass[10pt,twocolumn,letterpaper]{article}

\usepackage{cvpr}
\usepackage{times}
\usepackage{epsfig}
\usepackage{float}
\usepackage{graphicx}
\usepackage{amsmath}
\usepackage{amssymb}
\usepackage{subcaption}

\usepackage[pagebackref=true,breaklinks=true,letterpaper=true,colorlinks,bookmarks=false]{hyperref}

\cvprfinalcopy 

\def\cvprPaperID{12} 

\ifcvprfinal\fi
\begin{document}

\title{Fooling automated surveillance cameras: \\ adversarial patches to attack person detection}

\author{Simen Thys$^*$\\
{\tt\small simen.thys@student.kuleuven.be}\
\and
Wiebe Van Ranst$^*$\\
{\tt\small wiebe.vanranst@kuleuven.be}\
\and
Toon Goedem\'e\\
{\tt\small toon.goedeme@kuleuven.be}\\
KU Leuven\\
EAVISE, Technology Campus De Nayer, KU Leuven, Belgium.\\
$^*$ Authors contributed equally to this paper.
}

\maketitle

\begin{abstract}
Adversarial attacks on machine learning models have seen increasing interest in the past years. By making only  subtle changes to the input of a convolutional neural network, the output of the network can be swayed to output a completely different result.  The first attacks did this by changing pixel values of an input image slightly to fool a classifier to output the wrong class. Other approaches have tried to learn ``patches'' that can be applied to an object to fool detectors and classifiers. Some of these approaches have also shown that these attacks are feasible in the real-world, i.e. by modifying an object and filming it with a video camera. However, all of these approaches target classes that contain almost no intra-class variety (e.g. stop signs). The known structure of the object is then used to generate an adversarial patch on top of it.

In this paper, we present an approach to generate adversarial patches to targets with lots of intra-class variety, namely persons. The goal is to generate a patch that is able successfully  hide a person from a person detector. An attack that could for instance be used maliciously to circumvent surveillance systems, intruders can sneak around undetected by holding a small cardboard plate in front of their body aimed towards the surveilance camera.

From our results we can see that our system is able significantly lower the accuracy of a person detector. Our approach also functions well in real-life scenarios where the patch is filmed by a camera.
To the best of our knowledge we are the first to attempt this kind of attack on targets with a high level of intra-class variety like persons.

\end{abstract}
\thispagestyle{empty}
\section{Introduction}\begin{figure}
    \centering
    \includegraphics[width=\columnwidth]{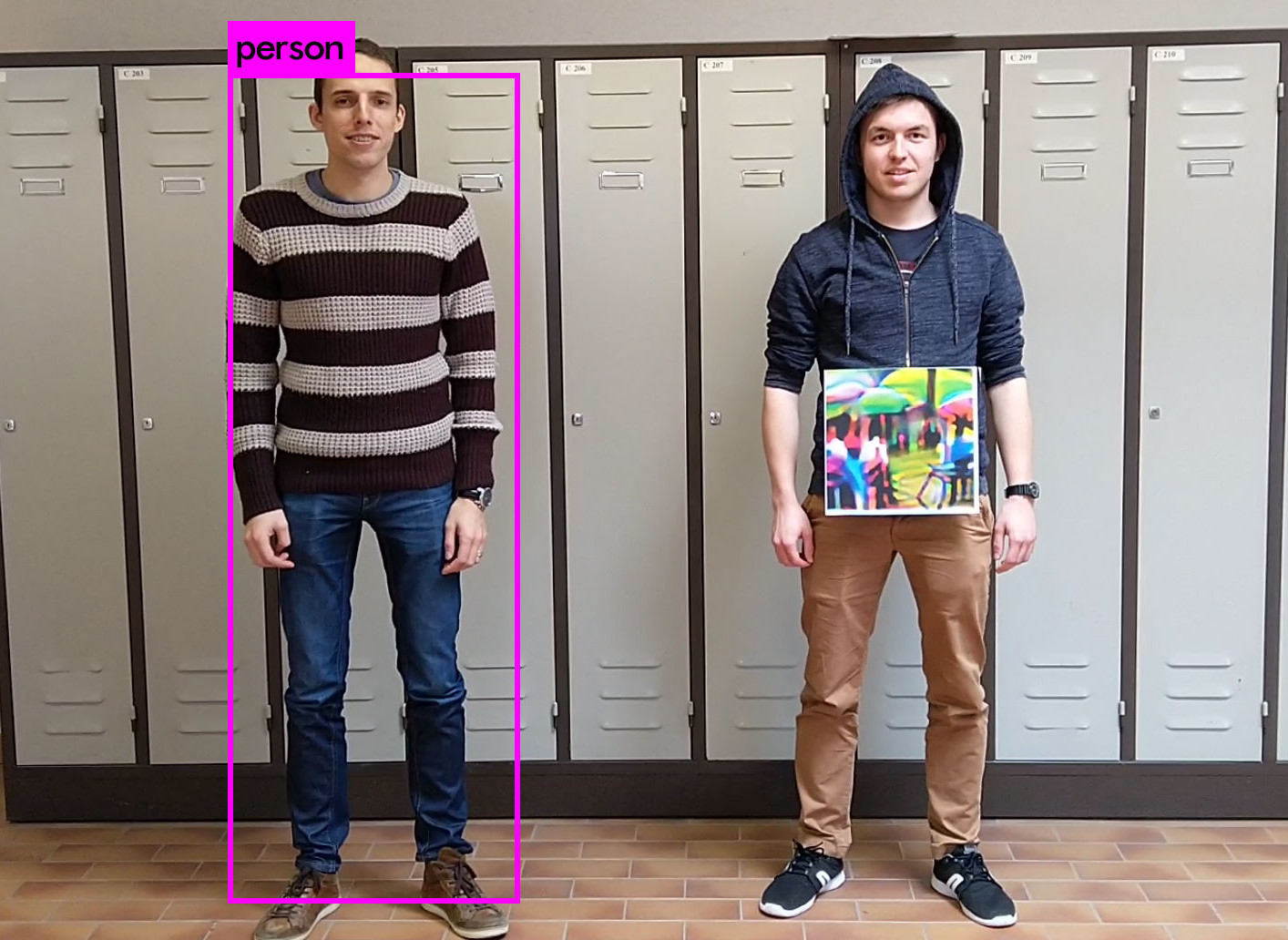}
    \caption{We create an adversarial patch that is successfully able to hide persons from a person detector. Left: The person without a patch is successfully detected. Right: The person holding the patch is ignored.}
    \label{fig:topfig}
\end{figure}
The rise of Convolutional Neural Networks (CNNs) has seen huge successes in the field of computer vision. The data-driven end-to-end pipeline in which CNNs learn on images has proven to get the best results in a wide range of computer vision tasks. Due to the depth of these architectures, neural networks are able to learn very basic filters at the bottom of the network (where the data comes in) to very abstract high level features at the top. To do this, a typical CNN contains millions of learned parameters. While this approach results in very accurate models, the interpretability decreases dramatically. Understanding exactly {\em why} a network classifies an image of a person as a person is very hard. The network has learned what a person looks likes by looking at many pictures of other persons. By evaluating the model we can determine how well the model work for person detection by comparing it to human annotated images. Evaluating the model in such a way however only tells us how well a detector performs on a certain test set. This test set does not typically contain examples that are designed to steer the model in the wrong way, nor does it contains examples that are especially targeted to fool the model. This is fine for applications where attacks are unlikely such as for instance fall detection for elderly people, but can pose a real issue in for instance security systems. A vulnerability in the person detection model of a security system might be used to circumvent a surveillance camera that is used for break in prevention in a building.

In this paper we highlight the risks of such an attack on person detection systems. We create a small (around $40 \text{cm} \times 40 \text{cm}$) ``adverserial patch'' that is used as a cloaking device to hide people from object detectors. A demonstration of this is shown in Figure~\ref{fig:topfig}.

The rest of this paper is structured as follows: Section~\ref{sec:rel_work} goes over the related work on adversarial attacks. Section~\ref{sec:adv_patches} discusses how we generate these patches. In Section~\ref{sec:results} we evaluate our patch both quantitatively on the Inria dataset, and qualitatively on real-life video footage taken while holding a patch. We reach a conclusion in Section~\ref{sec:conclusion}.

Source code is available at: \url{https://gitlab.com/EAVISE/adversarial-yolo}

\section{Related work}\label{sec:rel_work}
With the rise in popularity of CNNs, adversarial attacks on CNNs have seen an increase in popularity in the past years. In this section we go over the history of these kind of attacks. We first talk about digital attacks on classifiers, then talk about real-world attacks both for face recognition and object detection. Then we briefly discuss the object detector, YOLOv2 that in this work is the target of our attacks.

\paragraph{Adversarial attacks on classification tasks} Back in 2014  Bigio~et~at.~\cite{biggio2013evasion} showed the existence of adversarial attacks. After that, Szegedy~et~al.~\cite{szegedy2013intriguing} succeeded in generating adversarial attacks for classification models. They use a method that is able to fool the network to miss-classify an image, while only changing the pixel values of the image slightly so that the change is not visible to the human eye. Following that, Goodfellow~et~al.~\cite{goodfellow2014explaining} create a faster gradient sign method that made it more practical (faster) to generate adversarial attacks on images. Instead of finding the most optimal image as in~\cite{szegedy2013intriguing}, they find a single image in a larger set of images that is able to do an attack on the network. In~\cite{moosavi2016deepfool}, Moosavi-Dezfooli~et~al. present an algorithm that is able generate an attack by changing the image less and is also faster than the previous. They use hyper-planes to model the border between different output classes to the input image. Carlini~et~al.~\cite{carlini2017towards} present another adversarial attack, again, using optimisation methods, they improve in both accuracy and difference in images (using different norms) compared to the already mentioned attacks.
In~\cite{brown2017adversarial}~Brown~et~al. create a method that, instead of changing pixel values, generates patches that can be digitally placed on the image to fool a classifier. Instead of using one image, they use a variety of images to build in intra-class robustness.
In~\cite{evtimov2017robust} Evtimov~et~al. present a real-world attack for classification. They target the task of stop sign classification which proves to be challenging due to the different poses in which stop signs can occur. They generate a sticker than can be applied to a stop sign to make it unrecognizable. 
Athalye~et~al.~\cite{athalye2017synthesizing} present an approach in which the texture of a 3D model is optimized. Images of different poses are shown to the optimizer to build in robustness to different poses and lighting changes. The resulting object was then printed using a 3D printer.
The work of Moosavi-Dezfooli~\cite{moosavi2017universal} presents an approach to generate a single universal image that can be used as an adverserial perturbation on different images. The universal adversarial image is also shown to be robust to different detectors.

\paragraph{Real-world adversarial attack for face recognition}
An example of real-world adversarial attack is presented in~\cite{sharif2016accessorize}. Sharif~et~al. demonstrate the use of printed eyeglasses that can be used to fool facial recognition systems. To guarantee robustness the glasses need to work on a wide variety of different poses. To do this, they optimize the print on the glasses in such a way that they work on a large set of images instead of just a single image. They also include a Non Printability Score (NPS) which makes sure that the colors used in the image can be represented by a printer.

\paragraph{Real-world adversarial attacks for object detection}
Chen~et~al.~\cite{chen2018robust} present a real-world attack for object detection. They target the detection of stop signs in the Faster R-CNN detector~\cite{ren2015faster}. Like~\cite{athalye2017synthesizing}, they use the concept of Expectation over Transformation (EOT) (doing various transformation on the image) to build in robustness against different poses.
The most recent work we found to fool object detectors in the real-world is the work of Eykholt~et~al~\cite{song2018physical}. In it, they again target stop signs and use the YOLOv2~\cite{redmon2017yolo9000} detector to do a white box attack, where they fill in a pattern in the entire red area of the stop sign. They also evaluate on Faster-RCNN where they found that their attack also transfers to other detectors.

Compared to this work all attacks against object detectors focus on objects with fixed visual patterns like traffic signs and do not take into account intra-class variety. To the best of our knowledge no previous work has proposed a detection method that worked on a diverse class such as persons.

\paragraph{Object detection}
\begin{figure}
    \centering
    \includegraphics[width=\columnwidth]{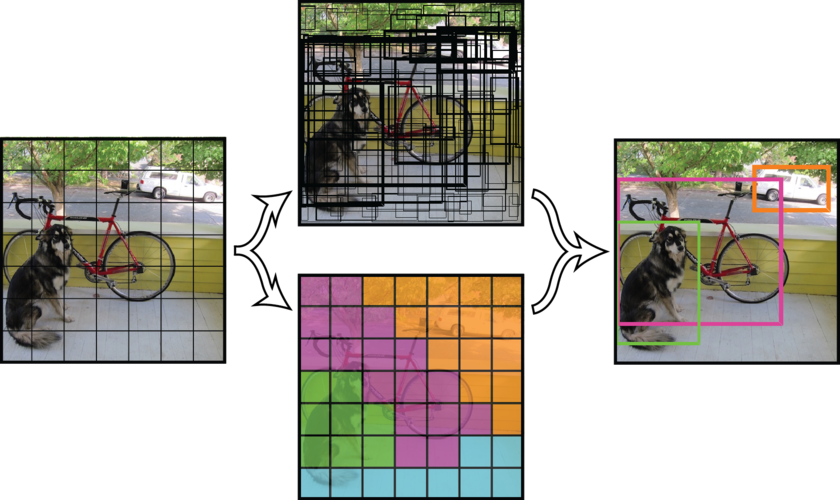}
    \caption{Overview of the YOLOv2 architecture. The detector outputs an objectness score (how likely it is that this detection contains an object), shown in the middle top figure, and a class score (which class is in the bounding box), shown in the middle bottom figure. Image source: \url{https://github.com/pjreddie/darknet/wiki/YOLO:-Real-Time-Object-Detection}}
    \label{fig:yolo}
\end{figure}
In this paper we target the popular YOLOv2~\cite{redmon2017yolo9000} object detector. YOLO fits in a bigger class of single shot object detectors (together with detectors like SSD~\cite{liu2016ssd}) where the bounding box, object score and class score is directly predicted by doing a single pass over the network. YOLOv2 is fully convolutional, an input image is passed to the network in which the various layers reduce it to an output grid with a resolution that is 32 times smaller than the original input resolution. Each cell in this output grid contains five predictions (called ``anchor points'') with bounding boxes containing different aspect ratios. Each anchor point contains a vector $[x_{\text{offset}}, y_{\text{offset}}, w, h, p_{\text{obj}}, p_{\text{cls}1}, p_{\text{cls}2}, ..., p_{\text{cls}n}]$. $x_\text{offset}$ and $y_\text{offset}$ is the position of the center of the bounding box compared to the current anchor point, $w$ and $h$ are the width and height of the bounding box, $p_\text{obj}$ is the probability that this anchor point contains an object, and $p_{\text{cls}1}$ through $p_{\text{clsn}}$ is the class score of the object learned using cross entropy loss. Figure~\ref{fig:yolo} shows an overview of this architecture.

\begin{figure*}
    \centering
    \includegraphics[width=0.8\textwidth]{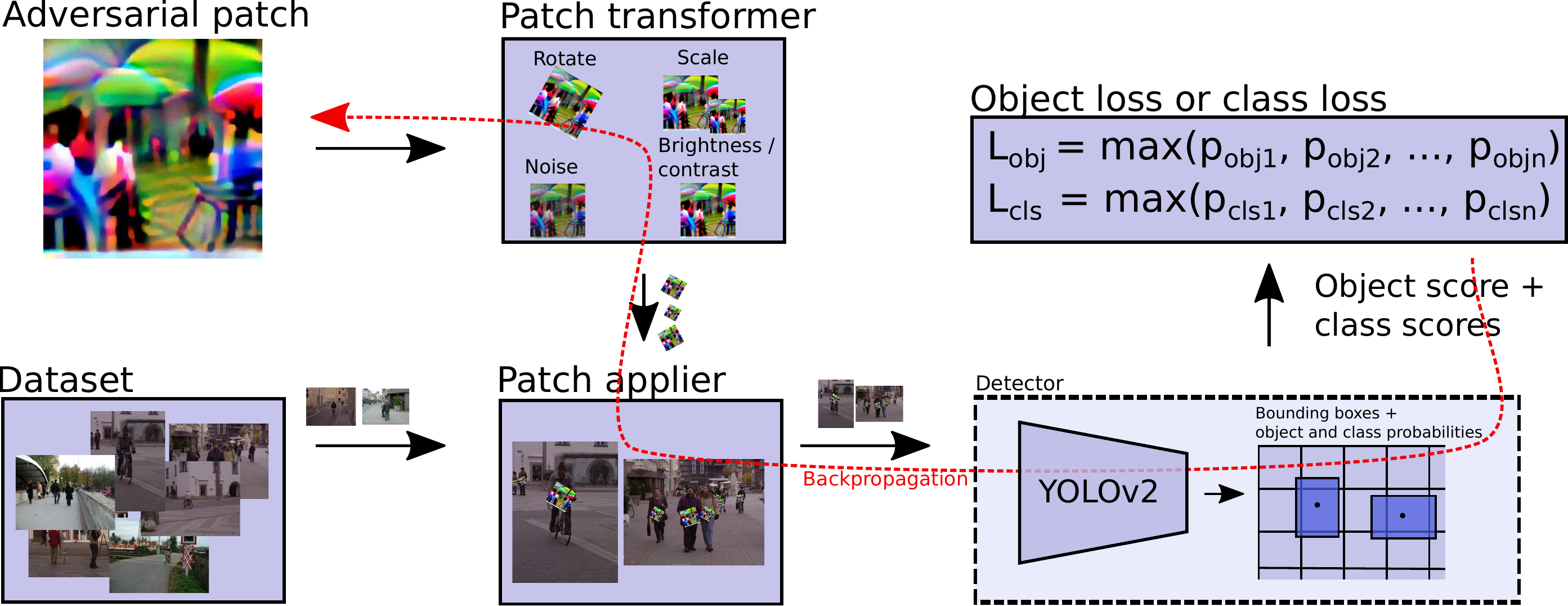}
    \caption{Overview of the pipeline to get the object loss.}
    \label{fig:adv_yolo}
\end{figure*}

\begin{figure}
    \centering
\begin{subfigure}[t]{.48\linewidth}
    \includegraphics[width=\columnwidth]{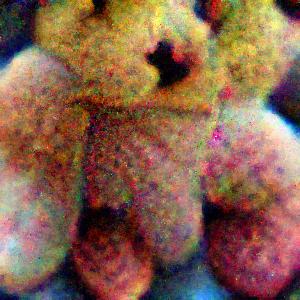}
    \caption{The resulting learned patch with an optimisation process that minimises classification and objectness score.}
    \label{fig:teddy_bear}
\end{subfigure}
\begin{subfigure}[t]{.48\linewidth}
    \includegraphics[width=\columnwidth]{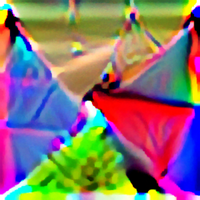}
    \caption{Another patch generated by minimising classification and detection score with slightly different parameters.}
    \label{fig:class_transfer}
\end{subfigure}
\begin{subfigure}[t]{.48\linewidth}
    \includegraphics[width=\columnwidth]{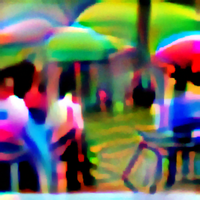}
    \caption{Patch generated by minimising the objectness score.}
    \label{fig:obj_patch}
\end{subfigure}
\begin{subfigure}[t]{.48\linewidth}
    \includegraphics[width=\columnwidth]{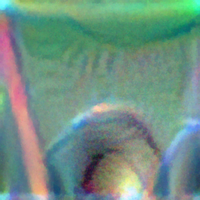}
    \caption{Minimising classification score only.}
    \label{fig:class_only}
\end{subfigure}
\caption{Examples of patches using different approaches.}
\label{fig:patches}
\end{figure}

\section{Generating adversarial patches against person detectors}\label{sec:adv_patches}
The goal of this work is create a system that is able to  generate printable adversarial patches that can be used to fool person detectors. As discussed earlier, Chen~et~al.~\cite{chen2018robust} and Eykholt~et~al.~\cite{song2018physical} already showed that adversarial attacks on object detectors in the real-world are possible. In their work they target stop signs, in this work we focus on persons which, unlike the uniform appearance of stop signs can vary a lot more. 
Using an optimisation process (on the image pixels) we try to find a patch that, on a large dataset, effectively lowers the accuracy of person detection.
In this section, we explain our process of generating these adversarial patches in depth.

Our optimisation goal consists of three parts:
\begin{itemize}
    \item $L_{nps}$ The non-printability score~\cite{sharif2016accessorize}, a factor that represents how well the colours in our patch can be represented by a common printer. Given by:
    $$
    L_{nps} = \sum_{p_\text{patch} \in p}\min_{c_\text{print} \in C}|p_\text{patch} - c_\text{print}|
    $$
    Where $p_\text{patch}$ is a pixel in of our patch $P$ and $c_\text{print}$ is a colour in a set of printable colours $C$. This loss favours colors in our image that lie closely to colours in our set of printable colours.
    \item $L_{tv}$ The total variation in the image as described in~\cite{sharif2016accessorize}. This loss makes sure that our optimiser favours an image with smooth colour transitions and prevents noisy images. We can calculate $L_{tv}$ from a patch $P$ as follows:
    $$
    L_{tv} = \sum_{i,j}\sqrt{((p_{i,j}-p_{i+1, j})^2 + (p_{i, j} - p_{i,j+1})^2}
    $$
    The score is low if neighbouring pixels are similar, and high if neighbouring pixel are different. 
    \item $L_\text{obj}$ The maximum objectness score in the image. The goal of our patch is to hide persons in the image. To do this, the goal of our training is to minimize the object or class score outputted by the detector. This score will be discussed in depth later in this section.
\end{itemize}
Out of these three parts follows our total loss function:
$$
L = \alpha L_{nps} + \beta L_{tv} + L_{obj}
$$
We take the sum of the three losses scaled by factors $\alpha$ and $\beta$ which are determined empirically, and optimise using the Adam~\cite{kingma2014adam} algorithm.

The goal of our optimizer is to minimise the total loss $L$. During the optimisation process we freeze all weights in the network, and change only the values in the patch. The patch is initialised on random values at the beginning of the process.

Figure~\ref{fig:adv_yolo} gives an overview of how the object loss is calculated. The same procedure is followed to calculate the class probability. In the remaining parts of this section we will explain how this is done in depth.

\subsection{Minimizing probability in the output of the detector}
As was explained in Section~\ref{sec:rel_work}, the YOLOv2 object detector outputs a grid of cells each containing a series of anchor points (five by default). Each anchor point contains the position of the bounding box, an object probability and a class score. To get the detector to ignore persons we experiment with three different approaches: We can either minimize the classification probability of class {\em person} (example patch in Figure~\ref{fig:class_only}, minimize the objectness score (Figure~\ref{fig:obj_patch}), or a combination of both (Figures~\ref{fig:class_transfer} and~\ref{fig:teddy_bear}). We tried out all approaches. Minimizing the class score has a tendency to switch the class person over to a different class. In our experiments with the YOLO detector trained on the MS COCO dataset~\cite{lin2014microsoft}, we found that the generated patch is detected as another class in the COCO dataset. Figure~\ref{fig:teddy_bear} and \ref{fig:class_transfer} is an example of taking the procuct of class and object probability, in the case of Figure~\ref{fig:teddy_bear},
the learned patch  ended up resembling a teddy bear, which it visually also resembles. The class ``'teddy bear' seemed to overpower the class ``person''. Because the patch starts to resemble another class however, the patch is less transferable to other models trained on datasets which do not contain the class.

The other approach we propose of minimising the objectness score does not have this issue. Although we only put it on top of people during the optimisation process, the resulting patch is less specific for a certain class than the other approach. Figure~\ref{fig:obj_patch} shows an example of such a patch.


\subsection{Preparing training data}
Compared to previous work done on stop signs~\cite{chen2018robust, song2018physical}, creating adversarial patches for the class persons is much more challenging:
\begin{itemize}
    \item The appearance of people varies much more: clothes, skin color, sizes, poses... Compared to stop signs which always have the same octagonal shape, and are usually red.
    \item People can appear in many different contexts. Stop signs mostly appear in the same context at the side of a street.
    \item The appearance of a person will be different depending on whether a person is facing away or towards the camera.
    \item There is no consistent spot on a person where we can put our patch. On a stop sign it's easy to calculate the exact position of a patch.
\end{itemize}

In this section we will explain how we deal with these challenges. Firstly, instead of artificially modifying a single image of the target object and doing different transformations as was done in~\cite{chen2018robust, song2018physical}, we use real images of different people. Our workflow is as follows: We first run the target person detector over our dataset of images. This yields bounding boxes that show where people occur in the image according to the detector. On a fixed position relative to these bounding boxes, we then apply the current version of our patch to the image under different transformations (which are explained in Section~\ref{sec:transforming}). The resulting image is then fed (in a batch together with other images) into the detector.
We measure the score of the persons that are still detected, which we use to calculate a loss function. Using back propagation over the entire network, the optimiser then changes the pixels in the patch further in order to fool the detector even more.

An interesting side effect of this approach is that we are not limited to annotated datasets. Any video or image collection can be fed into the target detector to generate bounding boxes. This allows our system to also do more targeted attacks. When we have data available from the environment we are targeting we can simply use that footage to generate a patch specific to that scene. Which will presumably preform better than a generic dataset.

In our tests we use the images of the Inria~\cite{dalal2005histograms} dataset.
These images are targeted more towards full body pedestrians which are better suited for our surveillance camera application.
We acknowledge that more challenging datasets like MS COCO~\cite{lin2014microsoft} and Pascal VOC~\cite{everingham2010pascal} are available, but they contain too much variety in which people occur (a hand is for instance annotated as person), making it hard to put our patch in a consistent position.

\subsection{Making patches more robust}\label{sec:transforming}
In this paper we target patches that have to be used in the real-world. This means that they are first printed out, and then filmed by a video camera. A lot of factors influence the appearance of the patch when  you do this: The lighting can change, the patch may be rotated slightly, the size of the patch with respect to the person can change, the camera may add noise or blur the patch slightly, viewing angles might be different\ldots To take this into account as much as possible, we do some transformations on the patch before applying it to the image. We do the following random transformations:
\begin{itemize}
    \item The patch is rotated up to 20 degrees each way.
    \item The patch is scaled up and down randomly
    \item Random noise is put on top of the patch.
    \item The brightness and contrast of the patch is changed randomly
\end{itemize}

Through this entire process it is important to note that it has to remain possible to calculate a backwards gradient on all operations all the way towards the patch.

\section{Results}\label{sec:results}

\begin{figure}
    \centering
    \includegraphics[width=\columnwidth]{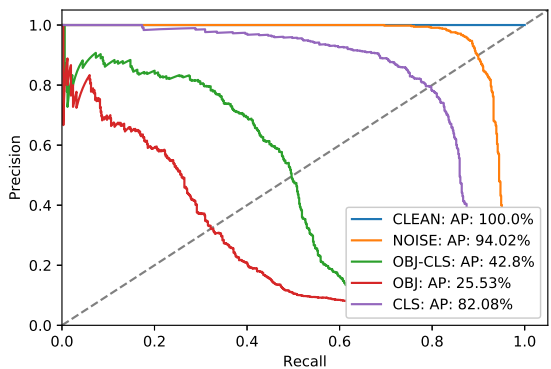}
    \caption{PR-curve of our different approaches (\texttt{OBJ-CLS}, \texttt{OBJ} and \texttt{CLS}), compared to a random patch (\texttt{NOISE}) and the original images (\texttt{CLEAN}).}
    \label{fig:pr_curve}
\end{figure}
\begin{figure*}
\centering
\begin{subfigure}[b]{.15\linewidth}
\includegraphics[width=\linewidth]{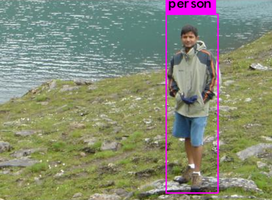}
\end{subfigure}
\begin{subfigure}[b]{.15\linewidth}
\includegraphics[width=\linewidth]{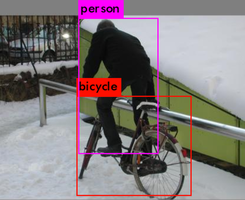}
\end{subfigure}
\begin{subfigure}[b]{.15\linewidth}
\includegraphics[width=\linewidth]{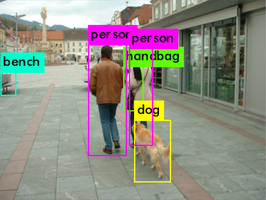}
\end{subfigure}
\begin{subfigure}[b]{.15\linewidth}
\includegraphics[width=\linewidth]{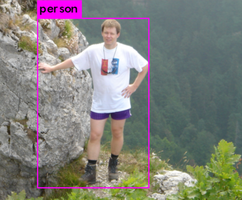}
\end{subfigure}
\begin{subfigure}[b]{.15\linewidth}
\includegraphics[width=\linewidth]{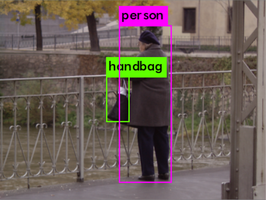}
\end{subfigure}
\begin{subfigure}[b]{.1\linewidth}
\includegraphics[width=\linewidth]{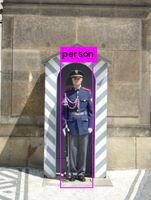}
\end{subfigure}

\centering
\begin{subfigure}[b]{.15\linewidth}
\includegraphics[width=\linewidth]{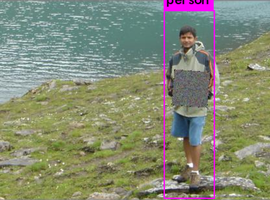}
\end{subfigure}
\begin{subfigure}[b]{.15\linewidth}
\includegraphics[width=\linewidth]{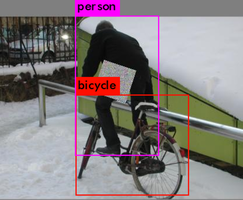}
\end{subfigure}
\begin{subfigure}[b]{.15\linewidth}
\includegraphics[width=\linewidth]{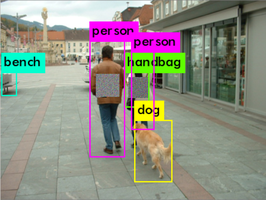}
\end{subfigure}
\begin{subfigure}[b]{.15\linewidth}
\includegraphics[width=\linewidth]{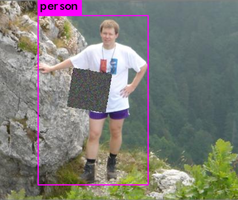}
\end{subfigure}
\begin{subfigure}[b]{.15\linewidth}
\includegraphics[width=\linewidth]{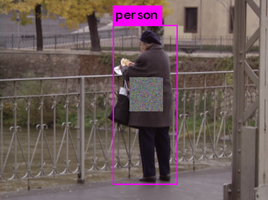}
\end{subfigure}
\begin{subfigure}[b]{.1\linewidth}
\includegraphics[width=\linewidth]{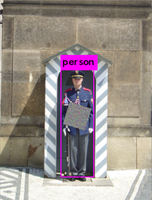}
\end{subfigure}

\centering
\begin{subfigure}[b]{.15\linewidth}
\includegraphics[width=\linewidth]{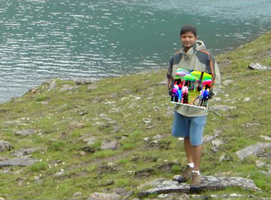}
\end{subfigure}
\begin{subfigure}[b]{.15\linewidth}
\includegraphics[width=\linewidth]{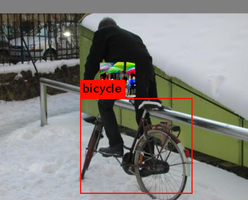}
\end{subfigure}
\begin{subfigure}[b]{.15\linewidth}
\includegraphics[width=\linewidth]{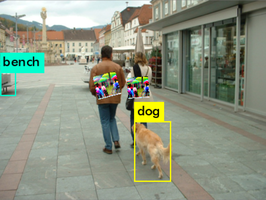}
\end{subfigure}
\begin{subfigure}[b]{.15\linewidth}
\includegraphics[width=\linewidth]{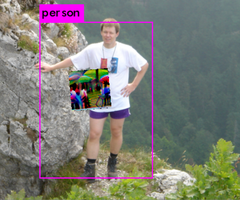}
\end{subfigure}
\begin{subfigure}[b]{.15\linewidth}
\includegraphics[width=\linewidth]{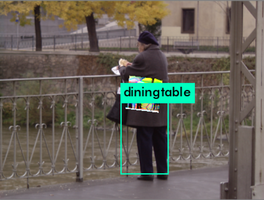}
\end{subfigure}
\begin{subfigure}[b]{.1\linewidth}
\includegraphics[width=\linewidth]{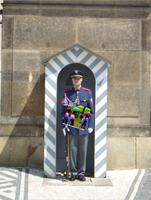}
\end{subfigure}

\caption{Examples of our output on the Inria testset.}
\label{fig:examples}
\end{figure*}

In this section we evaluate the effectiveness of our patches. We evaluate our patches by applying them to the Inria test set using the same process we used during training, including random transformations. In our experiments we tried to minimise a few different parameters that have the potential to hide persons. As a control, we also compare our results to a patch containing random noise that was evaluated in the exact same way as our random patches. Figure~\ref{fig:pr_curve} shows the result of our different patches. The objective in \texttt{OBJ-CLS} was to minimise the product of the object score and the class score, in \texttt{OBJ} only the object score, and in \texttt{CLS} only the class score. \texttt{NOISE} is our control patch of random noise, and \texttt{CLEAN} is the baseline with no patch applied. (Because the bounding boxes where generated by running the same detector over the dataset we get a perfect result.) From this PR-curve we can clearly see the impact a generated patch (\texttt{OBJ-CLS}, \texttt{OBJ} and \texttt{CLS}) has compared to a random patch which acts as a control. We can also see that minimising the object score (\texttt{OBJ}) has the biggest impact (lowest Average Precision (AP)) compared to using the class score.

A typical way to determine a good working point on a PR-curve to use for detecton is to draw a diagonal line on the PR-curve (dashed line in Figure~\ref{fig:pr_curve}), and look where it intersects with the PR-curve. If we do this for the \texttt{CLEAN} PR-curve, we can use the resulting threshold at that working point (0.4 in our case) as a reference to see how much our approach would lower the recall of the detector. In other words we ask the question: How many of the alarms generated by a surveillance system are circumvented by using our patches?
Table~\ref{tab:recall} shows the result of this analysis using abbreviations from Figure~\ref{fig:pr_curve}. From this we can clearly see that using our patch (\texttt{OBJ-CLS}, \texttt{OBJ} and \texttt{CLS}) significantly lowers the amount of generated alarms.

\begin{table}[]
    \centering
    \begin{tabular}{l | c}
         \textbf{Approach} & \textbf{Recall} (\%)\\
         \hline
         \texttt{CLEAN} &  100\\
         \texttt{NOISE} &  87.14\\
         \texttt{OBJ-CLS} & 39.31\\
         \texttt{OBJ} &  \textbf{26.46}\\
         \texttt{CLS} &  77.58\\
    \end{tabular}
    \caption{Comparison of different approaches in recall. How well do different approaches circumvent alarms?}
    \label{tab:recall}
\end{table}

Figure~\ref{fig:examples} shows examples of the patch applied to some images in the Inria test set. We apply the YOLOv2 detector first on images without a patch (row 1), with a random patch (row 2) and with our best generated patch which is \texttt{OBJ} (row 3). In most cases our patch is able to successfully hide the person from the detector. Where this is not the case, the patch is not aligned to the center of the person. Which can be explained by the fact that, during optimisation, the patch is also only positioned in the center of the person determined by the bounding box.

\begin{figure*}
\centering
\begin{subfigure}[b]{.24\linewidth}
\includegraphics[width=\linewidth]{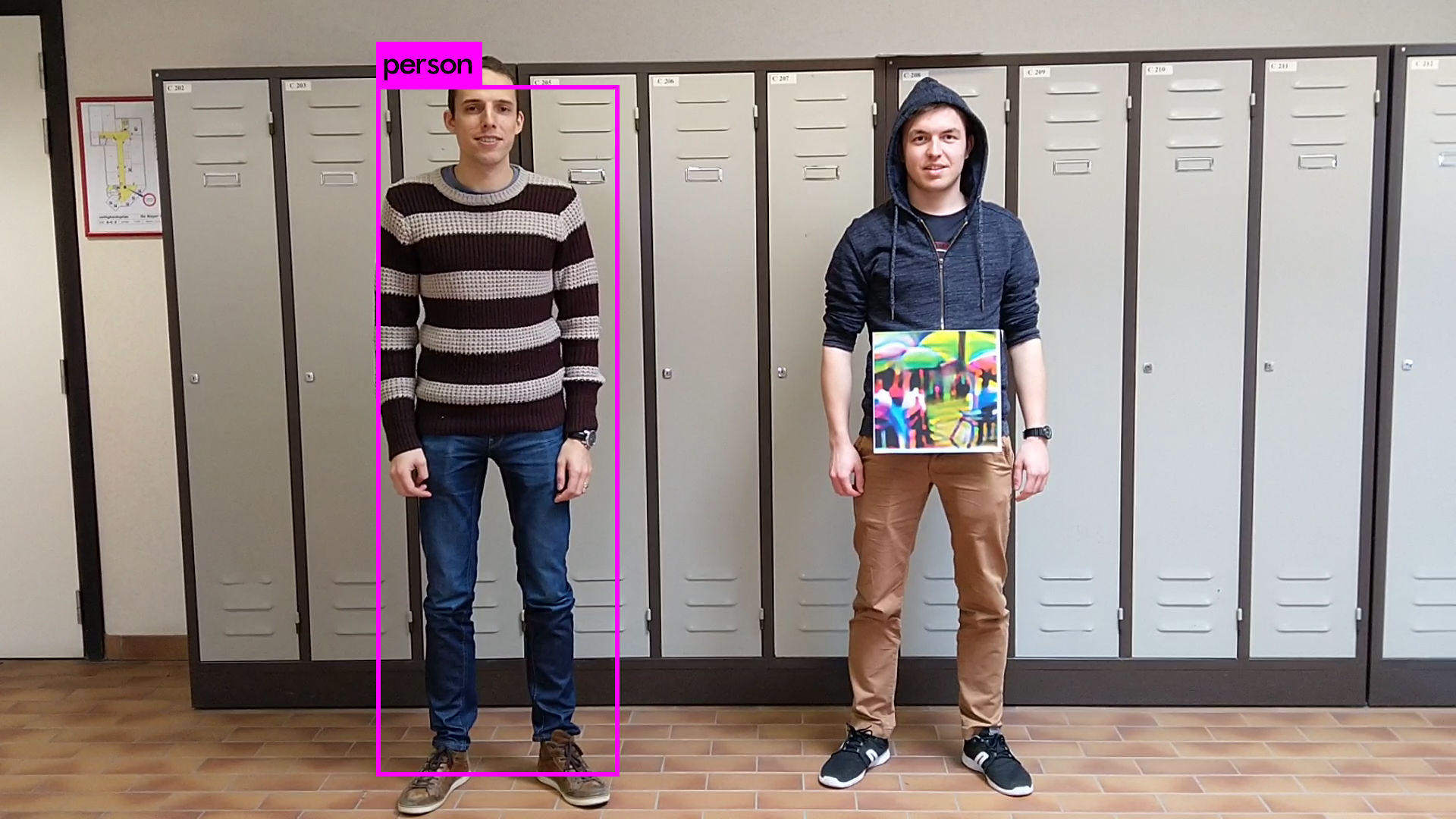}
\end{subfigure}
\begin{subfigure}[b]{.24\linewidth}
\includegraphics[width=\linewidth]{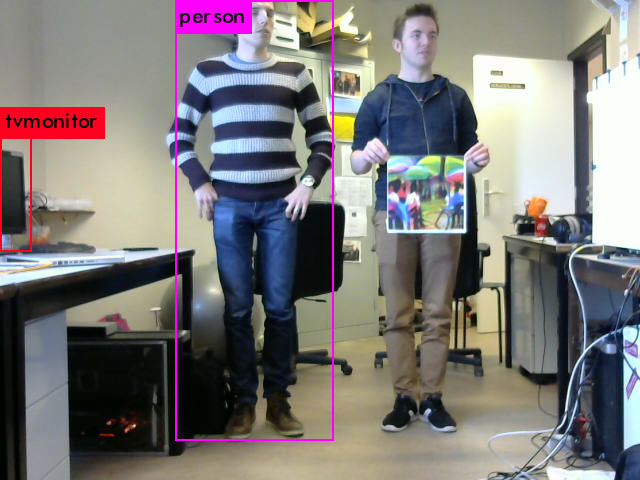}
\end{subfigure}
\begin{subfigure}[b]{.24\linewidth}
\includegraphics[width=\linewidth]{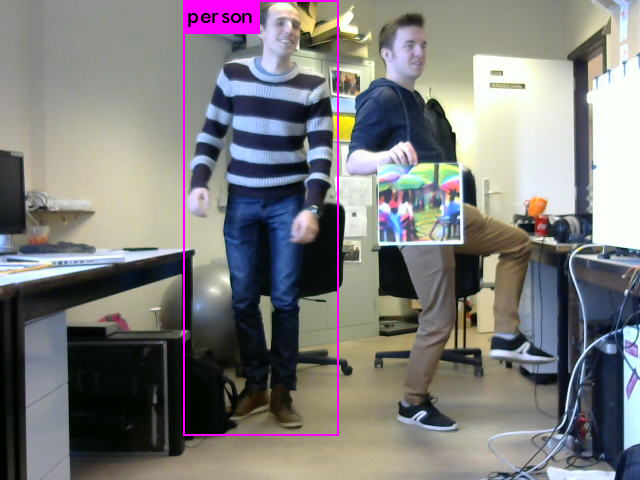}
\end{subfigure}
\begin{subfigure}[b]{.24\linewidth}
\includegraphics[width=\linewidth]{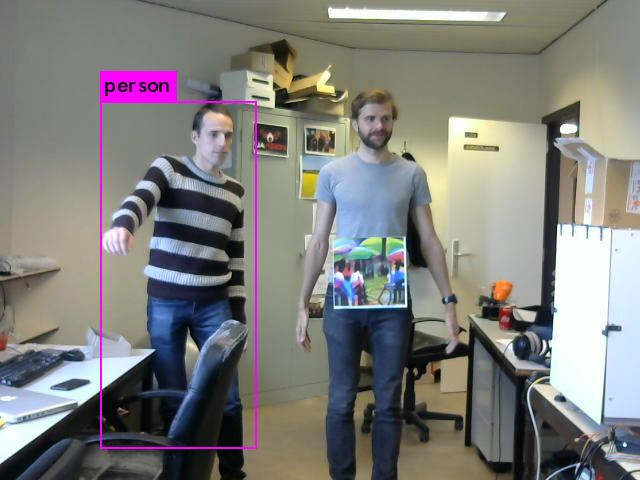}
\end{subfigure}

\begin{subfigure}[b]{.24\linewidth}
\includegraphics[width=\linewidth]{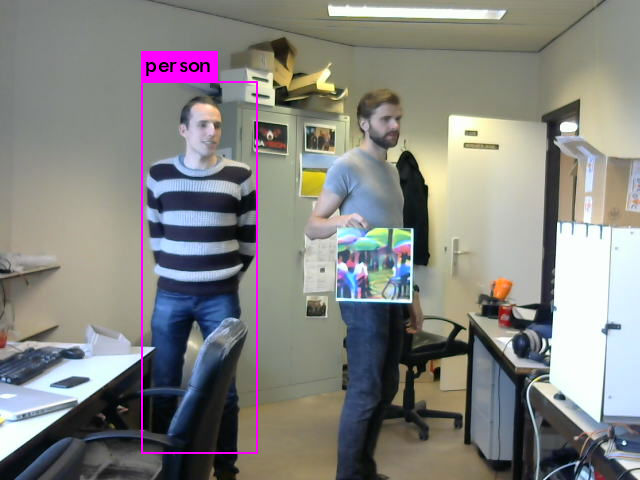}
\end{subfigure}
\begin{subfigure}[b]{.24\linewidth}
\includegraphics[width=\linewidth]{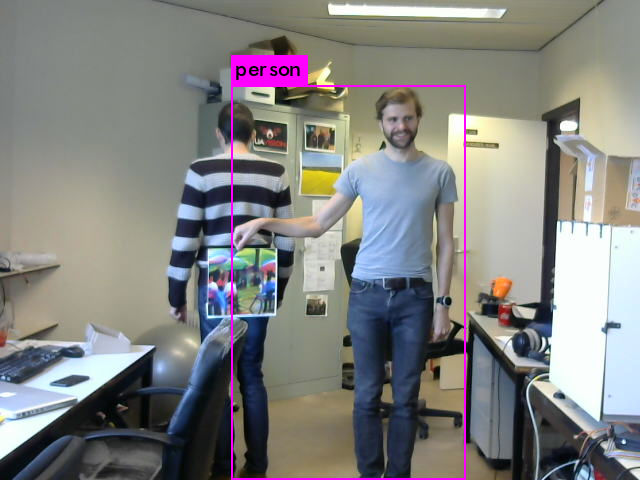}
\end{subfigure}
\begin{subfigure}[b]{.24\linewidth}
\includegraphics[width=\linewidth]{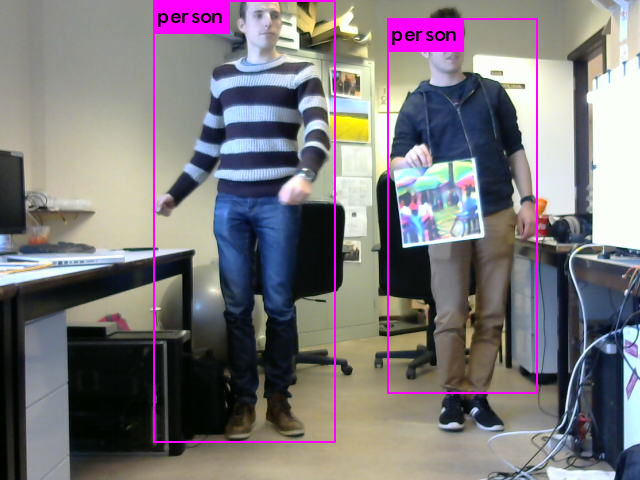}
\end{subfigure}
\begin{subfigure}[b]{.24\linewidth}
\includegraphics[width=\linewidth]{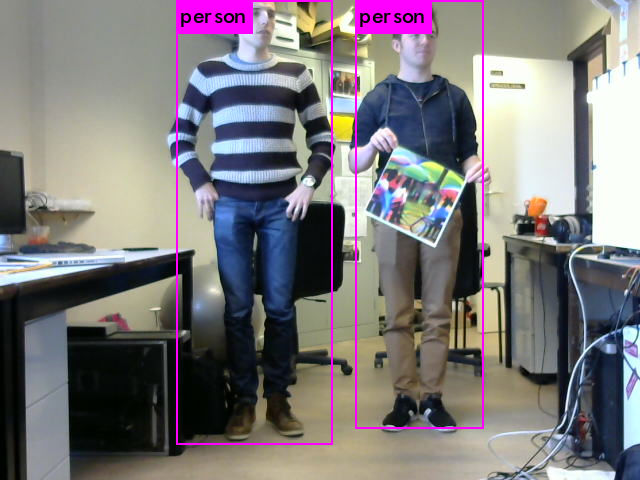}
\end{subfigure}

\caption{Real-world footage using a printed version of our patch.}
\label{fig:real_examples}
\end{figure*}

In Figure~\ref{fig:real_examples} we test how well a printed version of our patch works in the real world. In general the patch seems to work quite well. Due to the fact that the patch is trained on a fixed position relative to the bounding box holding the patch on the correct position seems to be quite important. A demo video can be found at: \url{https://youtu.be/MIbFvK2S9g8}.

\section{Conclusion}\label{sec:conclusion}
In this paper, we presented a system to generate adversarial patches for person detectors that can be printed out and used in the real-world. We did this by optimising an image to minimise different probabilities related to the appearance of a person in the output of the detector. In our experiments we compared different approaches and found that minimising object loss created the most effective patches. 


From our real-world test with printed out patches we can also see that our patches work quite well in hiding persons from object detectors, suggesting that security systems using similar detectors might be vulnerable to this kind of attack.

We believe that, if we combine this technique with a sophisticated clothing simulation, we can design a T-shirt print that can make a person virtually invisible for automatic surveillance cameras (using the YOLO detector).

\section{Future work}
In the future we would like to extend this work by making it more robust. One way to do this is by doing more (affine) transformation on the input data or using simulated data (i.e. apply the patch as a texture on a 3D-model of a person). 
Another area where more work can be done is transferability. Our current patches do not transfer well to completely different architectures like Faster R-CNN~\cite{ren2015faster}, optimising for different architectures at the same time might improve upon this.

{\small
\bibliographystyle{ieee}
\bibliography{egbib}
}

\end{document}